\documentclass[10pt,twocolumn,letterpaper]{article}
\pdfoutput=1
\usepackage{cvpr}              

\usepackage{graphicx}
\usepackage{amsmath}
\usepackage{amssymb}
\usepackage{booktabs}

\usepackage{balance}

\usepackage[pagebackref,breaklinks,colorlinks]{hyperref}

\usepackage[capitalize]{cleveref}
\crefname{section}{Sec.}{Secs.}
\Crefname{section}{Section}{Sections}
\Crefname{table}{Table}{Tables}
\crefname{table}{Tab.}{Tabs.}
\usepackage{verbatim}
\usepackage{hyperref}

\def\cvprPaperID{20} 
\def\confName{CVPR}
\def\confYear{2023}

\author{\textbf{Setareh Dabiri}$^{1}$ \qquad \textbf{Vasileios Lioutas}$^{1,2}$ \qquad \textbf{Berend Zwartsenberg}$^{1}$ \qquad \textbf{Yunpeng Liu}$^{1,2}$ \\
\textbf{Matthew Niedoba}$^{1,2}$ \qquad \textbf{Xiaoxuan Liang}$^{1,2}$ \qquad \textbf{Dylan Green}$^{1,2}$ \qquad \textbf{Justice Sefas}$^{1,2}$ \\
\textbf{Jonathan Wilder Lavington}$^{1,2}$ \qquad \textbf{Frank Wood}$^{1,2,3}$ \qquad \textbf{Adam \'Scibior}$^{1,2}$\\
$^1$\normalfont{Inverted AI}\qquad $^2$University of British Columbia\qquad $^3$Mila
}

\begin{document}

\title{Realistically distributing object placements in synthetic training data improves the performance of vision-based object detection models}
\maketitle



\begin{abstract}
    When training object detection models on synthetic data, it is important to make the distribution of synthetic data as close as possible to the distribution of real data.
    We investigate specifically the impact of object placement distribution, keeping all other aspects of synthetic data fixed. Our experiment, training a 3D vehicle detection model in CARLA and testing on KITTI, demonstrates a substantial improvement resulting from improving the object placement distribution.
\end{abstract}

\section{Introduction}
\label{sec:intro}

It is well known that any domain gap between training and test data hurts the performance of machine learning models in general, and object detectors in particular. When training with synthetic data obtained from simulations, the bulk of attention in the existing literature has been on the domain gap that has to do with the visuals, such as textures, lighting, weather, etc. (also referred to as the \emph{appearance gap}), while the impact of different types, numbers, and placements of objects (called the \emph{content gap}) has not been the primary area of research. In \cref{sec:related} we review existing work that addresses the content gap generally and the placement distribution in particular, but we believe the literature is lacking a clear demonstration of how much of an impact the placement distribution in synthetic data can have on the performance of vision-based object detectors in driving contexts. In this paper we test the hypothesis that the realism of physical object placement distribution in synthetic data has a significant impact on the performance of vision models trained on said data.

\begin{table}
  \centering
  \small
  \begin{tabular}{@{}lc@{}c@{}c@{}}
    \toprule
    &&AP11/AP40&\\
    Metric\hspace{12pt}Dataset & Easy & Moderate & Hard \\
    \midrule
    2D BBox Baseline & 56.7/56.9 & 42.1/40.6 & 35.2/32.8 \\
    \hspace{40pt}INITIALIZE& \textbf{67.3/67.7} & \textbf{51.3/49.9} & \textbf{43.7/40.9} \\
    \bottomrule
  \end{tabular}
  \caption{Average precision of 2D bounding boxes on KITTI validation set, predicted by models trained on synthetic datasets with baseline and realistic vehicle placements respectively. Predicted bounding boxes with IoU larger than 0.7 with ground truth are considered successful detections.}
  \label{tab:2Dbbox}
\end{table}

 \begin{figure*}[t]
  \centering
  \begin{subfigure}{0.47\linewidth}
    \includegraphics[width=\textwidth]{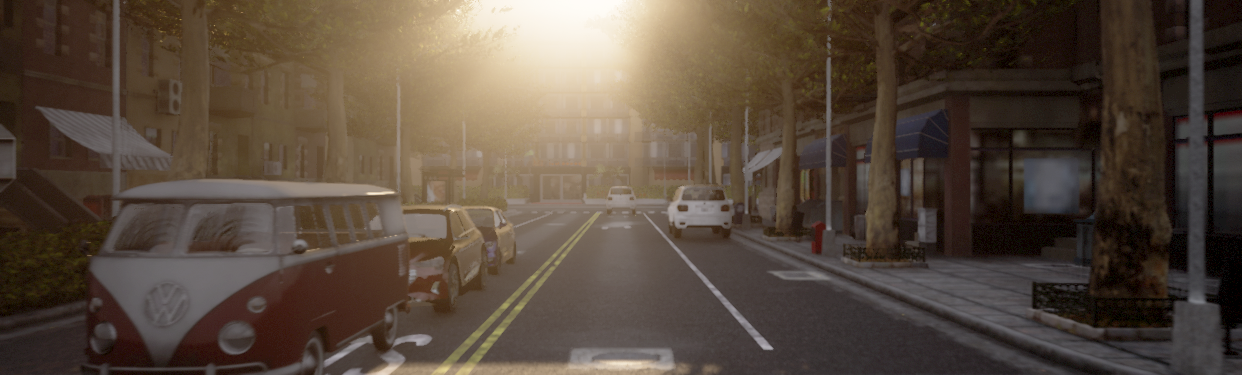}
  \end{subfigure}
  \begin{subfigure}{0.47\linewidth}
    \includegraphics[width=\textwidth]{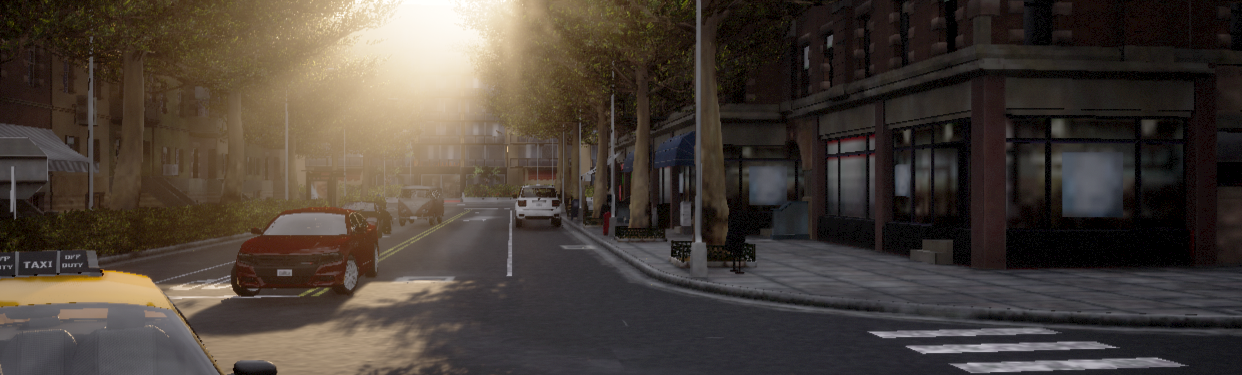}
  \end{subfigure}
  \begin{subfigure}{0.47\linewidth}
    \includegraphics[width=\textwidth]{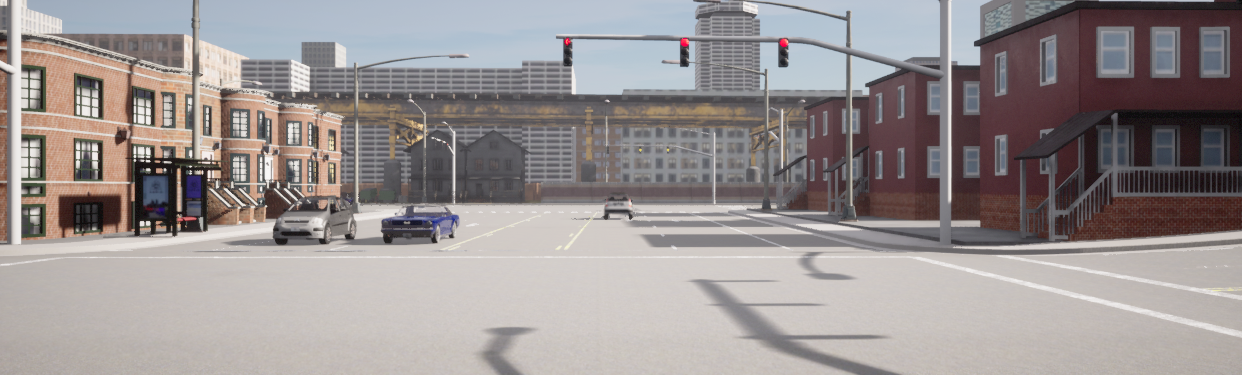}
  \end{subfigure}
  \begin{subfigure}{0.47\linewidth}
    \includegraphics[width=\textwidth]{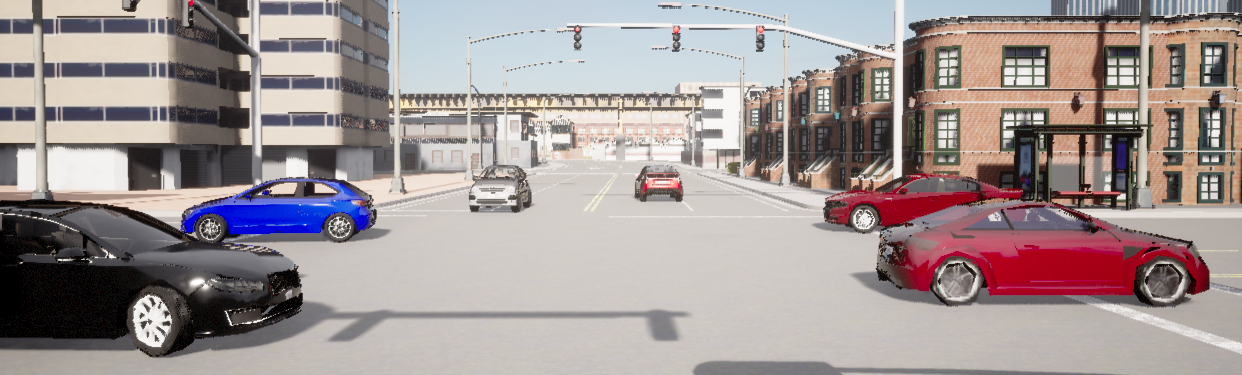}
  \end{subfigure}
  \begin{subfigure}{0.47\linewidth}
    \includegraphics[width=\textwidth]{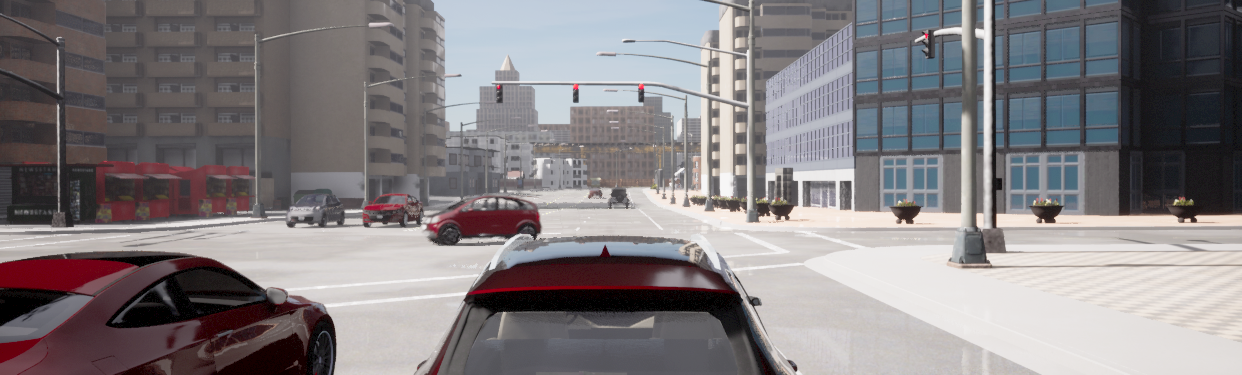}
  \end{subfigure}
  \begin{subfigure}{0.47\linewidth}
    \includegraphics[width=\textwidth]{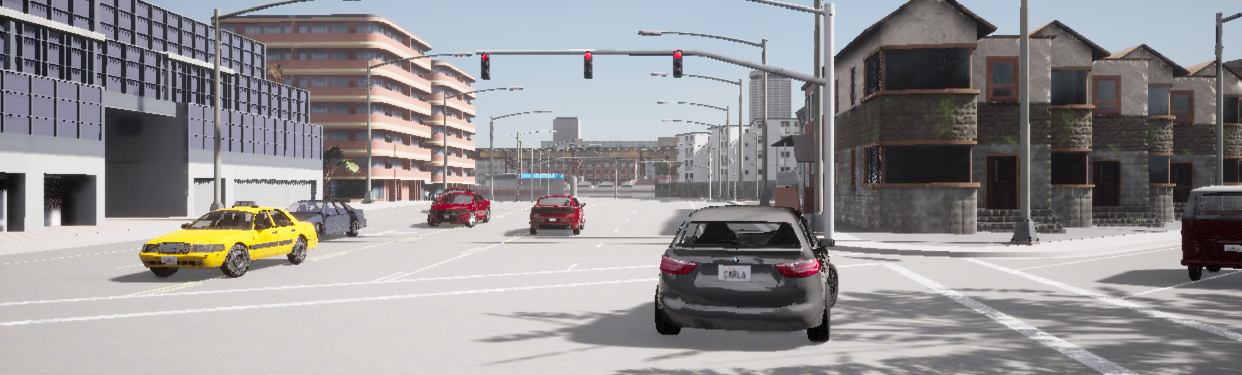}
  \end{subfigure}
  \begin{subfigure}{0.47\linewidth}
    \includegraphics[width=\textwidth]{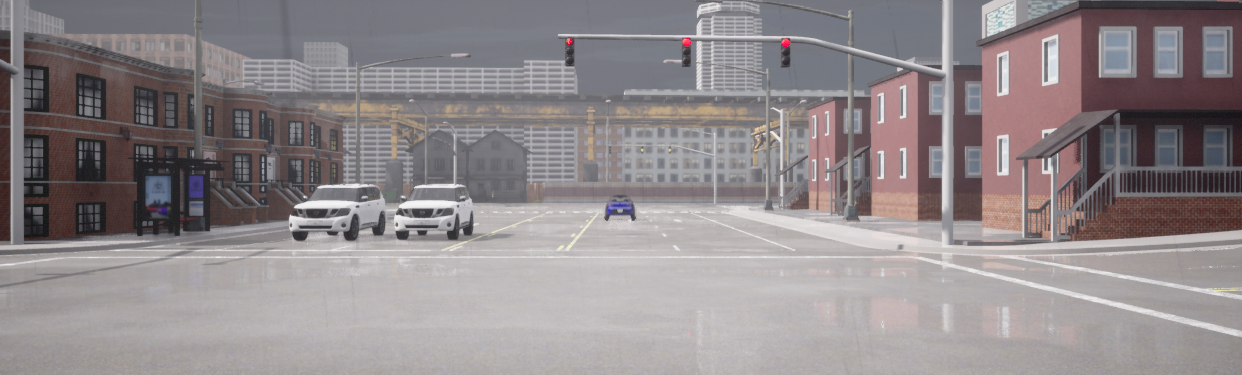}
  \end{subfigure}
  \begin{subfigure}{0.47\linewidth}
    \includegraphics[width=\textwidth]{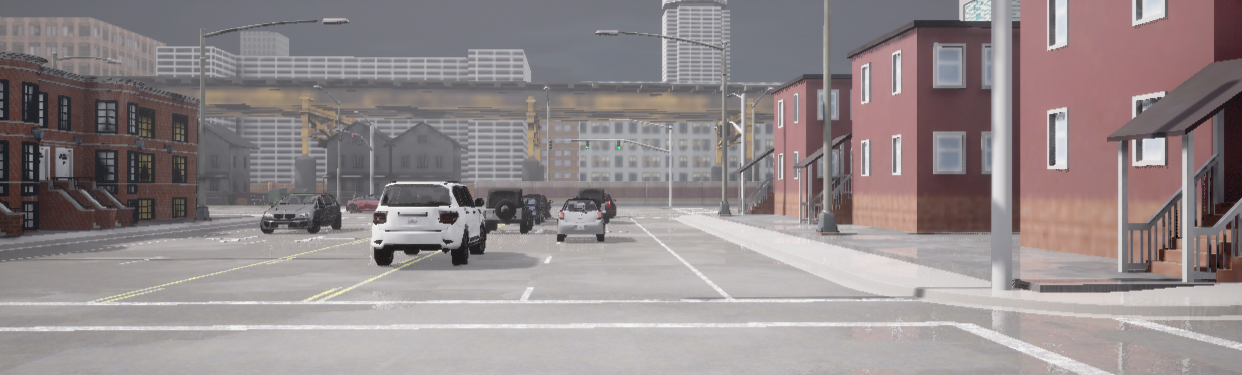}
  \end{subfigure}
  \begin{subfigure}{0.47\linewidth}
    \includegraphics[width=\textwidth]{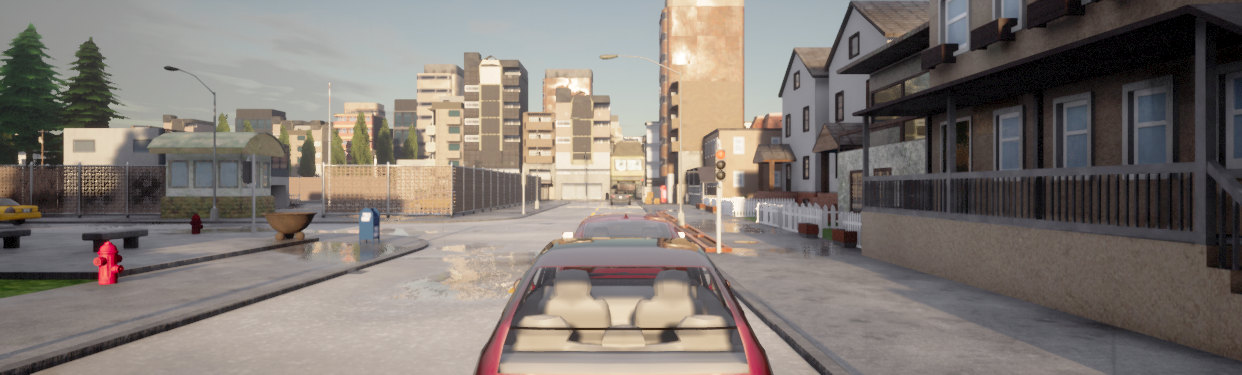}
    \caption{Baseline}
  \end{subfigure}
  \begin{subfigure}{0.47\linewidth}
    \includegraphics[width=\textwidth]{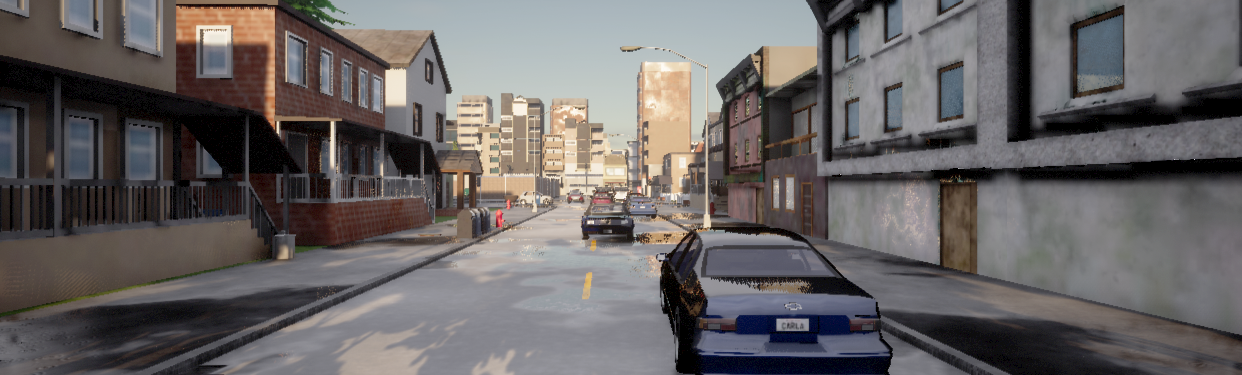}
    \caption{INITIALIZE}
  \end{subfigure}
  \caption{Sample training set images generated using CARLA. We compare the baseline placement (left) with a realistic one (right).}
  \label{fig:dataSample}
\end{figure*}

We use a carefully controlled experimental setup, where we generate training data using the CARLA driving simulator~\cite{dosovitskiy_carla_2017} and we use real validation data provided in KITTI~\cite{Geiger2012CVPR} as test data. We compare a \emph{baseline} object placement model, where we allow the CARLA Traffic Manager to freely move vehicles and take a snapshot of their positions at a particular time, with our commercial model\footnote{\url{https://docs.inverted.ai}}, called INITIALIZE, that jointly samples realistic vehicle placements. We specifically isolate the object placement distribution as the independent variable, fixing the object types, appearances, counts, as well as weather conditions and locations so that they match exactly between the two versions of the training dataset. Our results show a large improvement in test set performance arising from that single intervention.

We use a PGD~\cite{wang2021pgd} model for object detection. This choice is mostly orthogonal to the claims of the paper and we do not expect it to have a significant bearing on the results. We use the publicly available source code for PGD~\cite{mmdet3d2020}, and will release our training datasets and specific configurations for reproducibility purposes. We report standard quantitative performance metrics and provide some qualitative illustrations for the differences between the baseline and INITIALIZE training sets, as defined in the paragraph above, as well as the different test set predictions made by PGD trained on each of those datasets respectively. It is worth noting that we do not attempt to achieve competitive performance on KITTI, which would require addressing the appearance gap, as well as increasing the variety of vehicle models and locations, and perhaps additional training on real data. Instead, our focus is solely on isolating the impact of vehicle placement distribution in training data on test set performance.

\section{Experiments}

\subsection{Data generation}
We generate a KITTI-like dataset for 3D object detection from a forward-facing camera angle using CARLA. We manually designate regions of interest in Town 1, Town 2, Town 3, Town 4, and Town 10 to cover different road geometries. The baseline dataset uses the available "Traffic Manager" in CARLA to drive vehicles from predefined spawning locations, while the dataset with realistic vehicle placements samples vehicle positions directly based on traffic patterns learned from data.
In an attempt to obtain a dataset that is diverse in appearance, we generate scenes by varying the sun angle to simulate different times of day, as well as various weather conditions, including clear, cloudy, and rainy settings. To ensure fairness in the generation process for both datasets, we specify the same number of agents and episodes for each generated scene. Specifically, we place 15 vehicles within a circular region of interest (ROI) with a radius of 50 meteres. If the number of available CARLA spawn points within the ROI is less than 15, we lower the number of vehicles to match the number of spawn points available, in both versions of the dataset. This ensures that each image has the same number of vehicles within the same ROI in the same weather in both datasets. The number of vehicles visible in the image can vary between the datasets, since the camera does not capture all agents within the ROI.

We randomly assign one of the vehicles in the ROI as the ego vehicle and record snapshots from the camera located on the top of the ego agent for both datasets. To maintain consistency with the KITTI dataset, we save the images in 1242${\times}$375 resolution. Each dataset includes 1844 images. \Cref{fig:dataSample} depicts some samples from the baseline dataset and the INITIALIZE dataset. It is easy to see that the realistic placement produces more variability, in particular regarding the positioning of vehicles relative to the centerline.

\subsection{Results}
To demonstrate the importance of vehicle placement on realism in synthetically generated training data, we train a monocular 3D detection model named PGD~\cite{wang2021pgd}, using the source code provided by its original authors~\cite{mmdet3d2020}. We train two versions of this model, on baseline dataset and the dataset with realistic vehicle placements respectively, but otherwise identical, and evaluate the performance of both versions on the KITTI validation dataset, consisting of 3769 images. We use the same hyperparameters for both versions, the exact values of which will be released with the source code. The average precision of the 2D bounding box (2DBBox), bird's eye view (BEV), 3D bounding box (3D BBox) and average orientation similarity (AOS)~\cite{Geiger2012CVPR} of two trained models tested on the KITTI validation set are reported in tables \cref{tab:.0.7IoU}, \cref{tab:0.5IoU} and \cref{tab:2Dbbox}. 

\looseness=-1 The tables display the performance on three different difficulty levels defined in the KITTI dataset: Easy, Moderate, and Hard. \Cref{tab:.0.7IoU} presents results where bounding boxes with an overlap of more than 70\% are counted as positive, and \cref{tab:0.5IoU} displays the same for overlaps greater than 50\%. The criteria for determining the difficulty are the minimum bounding box height, maximum occlusion level, and maximum truncation percentage as described in \cite{Geiger2012CVPR}. The results include both object detection and orientation estimation, which are evaluated using the average precision (AP) and average orientation similarity (AOS) metrics, respectively. 
 
\begin{figure*}[t]
  \centering
    \begin{subfigure}{0.47\linewidth}
    \includegraphics[width=\textwidth]{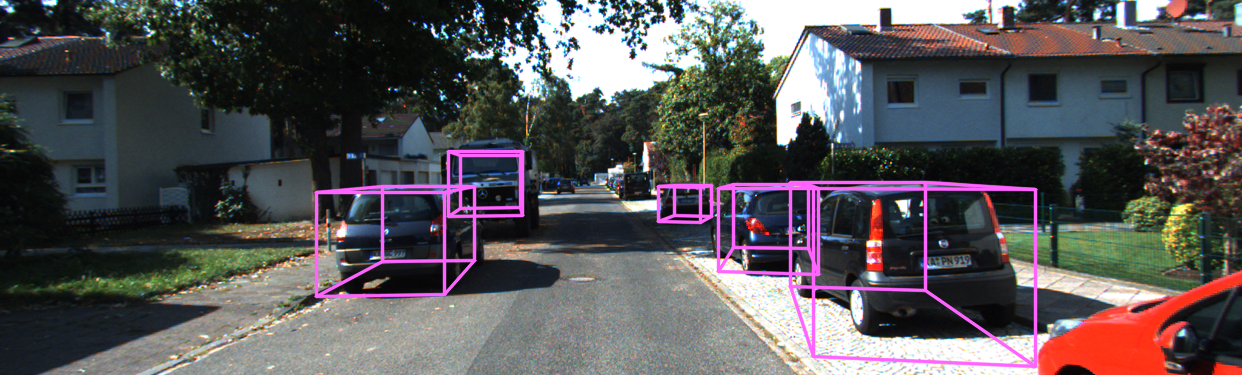}
  \end{subfigure}
  \begin{subfigure}{0.47\linewidth}
    \includegraphics[width=\textwidth]{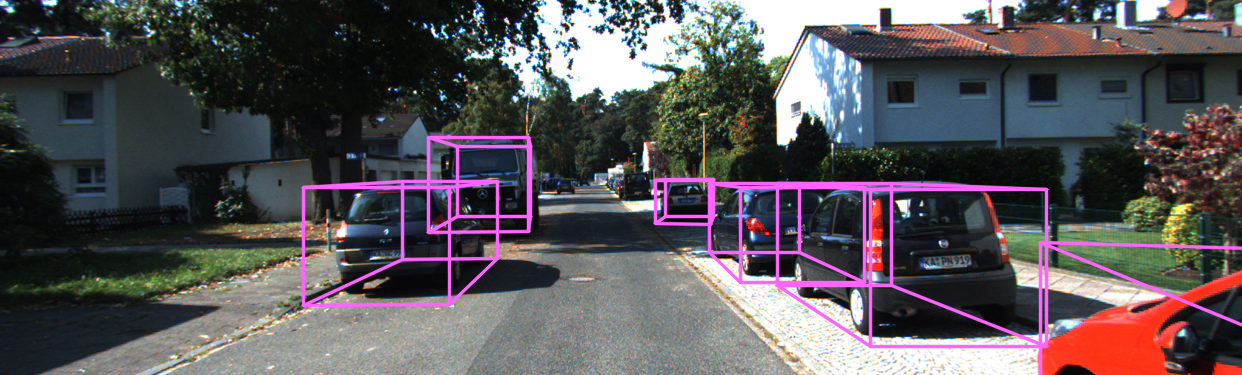}
  \end{subfigure}
    \begin{subfigure}{0.47\linewidth}
    \includegraphics[width=\textwidth]{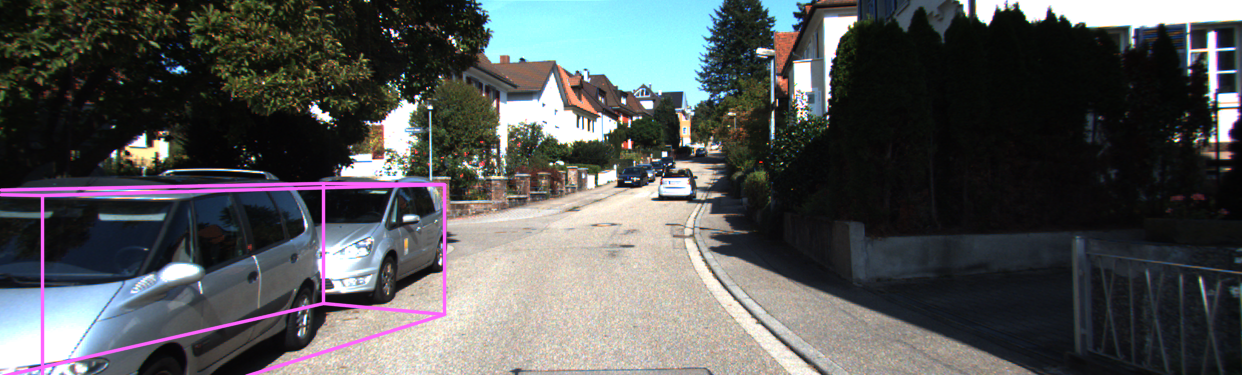}
  \end{subfigure}
  \begin{subfigure}{0.47\linewidth}
    \includegraphics[width=\textwidth]{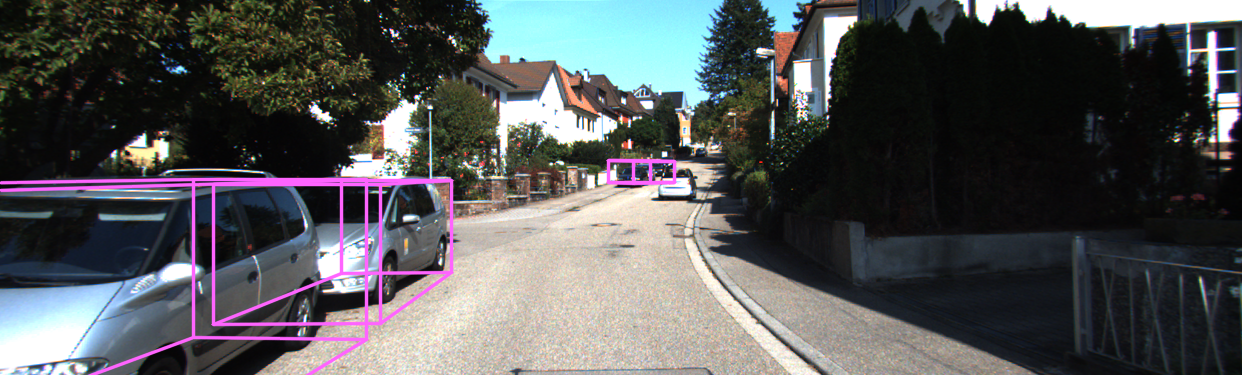}
  \end{subfigure}
        \begin{subfigure}{0.47\linewidth}
    \includegraphics[width=\textwidth]{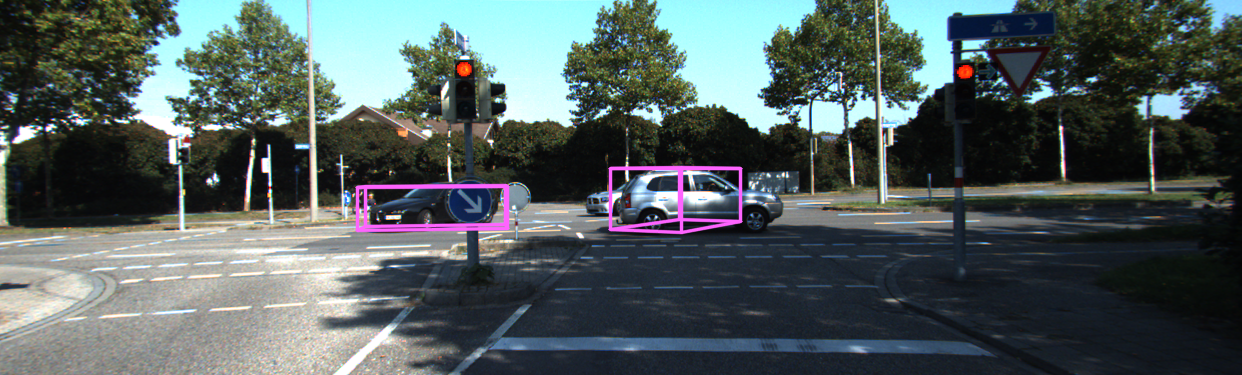}
  \end{subfigure}
  \begin{subfigure}{0.47\linewidth}
    \includegraphics[width=\textwidth]{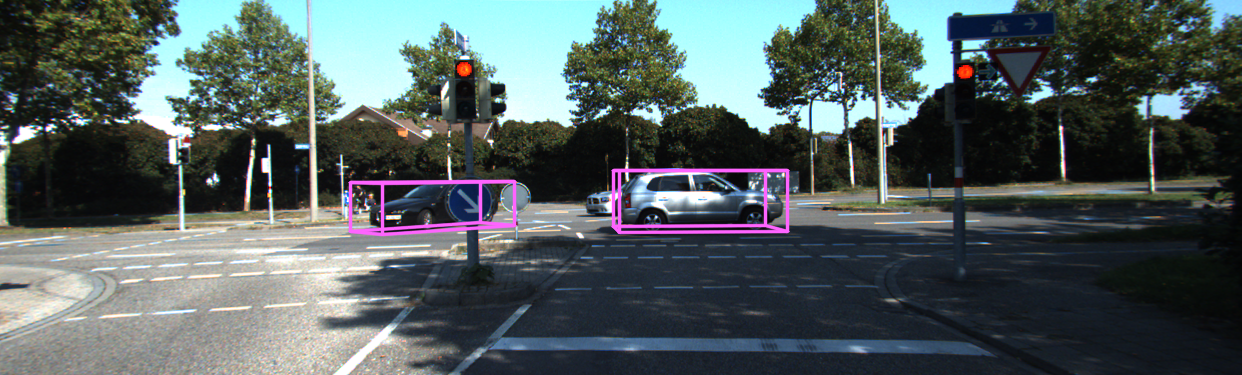}
  \end{subfigure}
  \begin{subfigure}{0.47\linewidth}
    \includegraphics[width=\textwidth]{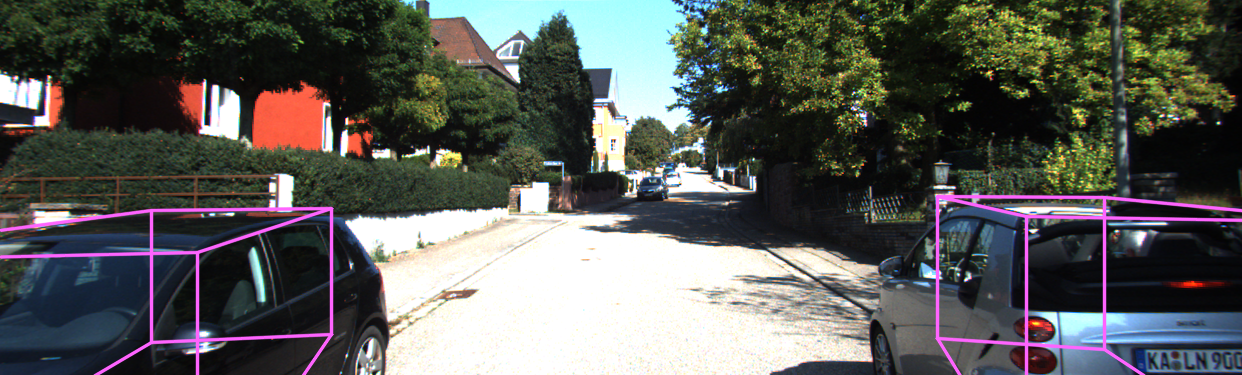}
    \caption{Baseline}
  \end{subfigure}
  \begin{subfigure}{0.47\linewidth}
    \includegraphics[width=\textwidth]{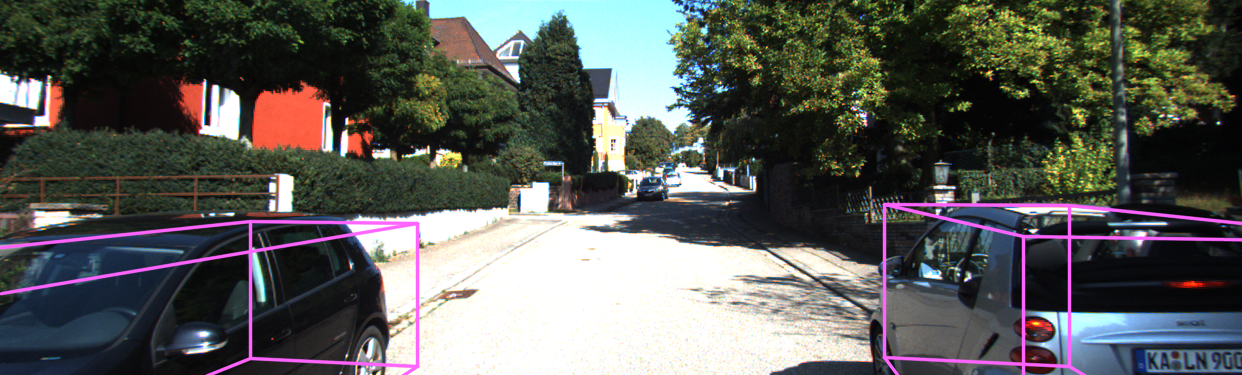}  
    \caption{INITIALIZE}
  \end{subfigure}
  \caption{3D bounding box predictions on the KITTI dataset. The left column depicts predictions of the model trained on synthetic data with baseline vehicle placements, while the right column shows predictions from the model trained on synthetic data with realistic vehicle placements.}
  \label{fig:resultsOnKitti}
\end{figure*}

 As evident from the data presented in \cref{tab:0.5IoU} and \cref{tab:.0.7IoU}, using realistic object placements drastically improves average precision of 3D bounding box and BEV of cars across all dataset difficulty categories. Moreover, as indicated in \cref{tab:.0.7IoU}, training the model on the dataset with realistic vehicle placements results in a considerable gain in the average orientation similarity of the predicted bounding boxes. \Cref{tab:2Dbbox} illustrates a substantial improvement in the average precision of 2D bounding boxes. \Cref{fig:resultsOnKitti} illustrates the predicted 3D bounding boxes on images from the KITTI validation set, once again showing that the realistic vehicle placement from INITIALIZE results in better performance on real data.  

\begin{table}
  \centering
  \small
  \begin{tabular}{@{}lc@{}c@{}c@{}}
    \toprule
    &&AP11/AP40&\\
    Metric\hspace{12pt}Dataset & Easy & Moderate & Hard \\
    \midrule
    BEV\hspace{20pt}Baseline & 0.68/0.35 & 0.56/0.2 & 0.53/0.19 \\
    \hspace{40pt}INITIALIZE &  \textbf{9.1/5.8} & \textbf{7.4/4.4} & \textbf{6.5/3.8} \\
    \midrule
    3D BBox Baseline& 0.32/0.13 & 0.32/0.10 & 0.32/0.04\\
    \hspace{40pt}INITIALIZE &  \textbf{6.8/2.8} & \textbf{5.8/2.2} & \textbf{5.6/1.9}\\
    \bottomrule
  \end{tabular}
  \caption{Average precision of BEV and 3D bounding boxes on KITTI validation set, predicted by models trained on synthetic datasets with baseline and realistic vehicle placements respectively. Predicted bounding boxes with IoU larger than 0.5 with ground truth are considered successful detections.}
  \label{tab:0.5IoU}
\end{table}

\begin{table}
  \centering
  \small
  \begin{tabular}{@{}lc@{}c@{}c@{}}
    \toprule
    &&AP11/AP40&\\
    Metric\hspace{12pt}Dataset & Easy & Moderate & Hard \\
    \midrule
    BEV\hspace{20pt}Baseline &  0.05/0.02 & 0.06/0.01 & 0.057/0.01\\
    \hspace{40pt}INITIALIZE  &  \textbf{0.27/0.11} & \textbf{1.51/0.09} & \textbf{1.51/0.03}\\
    \midrule
    3D BBox Baseline & 0.02/0.01 & 0.04/0.01 & 0.04/0.0\\
    \hspace{40pt}INITIALIZE & \textbf{0.07/0.01} & \textbf{1.51/0.01} & \textbf{1.51/0.01}\\
    \midrule
    AOS\hspace{20pt}Baseline  &17.7/17.9 & 13.7/13.2 & 14.1/11.2\\
    \hspace{40pt}INITIALIZE& \textbf{21.0/20.9} & \textbf{17.3/16.5} & \textbf{15.2/13.8}\\
    \bottomrule
  \end{tabular}
  \caption{Average precision of BEV, 3D bounding boxes and AOS on KITTI validation set, predicted by models trained on synthetic datasets with baseline and realistic vehicle placements respectively. Predicted bounding boxes with IoU larger than 0.7 with ground truth are considered successful detections.}
  \label{tab:.0.7IoU}
\end{table}

\section{Related work} \label{sec:related}

Training object detection models with synthetic data is a well-established approach, which is attractive mostly due to its relative low cost for both recording and annotating compared to data obtained in the real world. The literature on the subject is vast and the topic of many articles~\cite{peng_learning_2015,miao_parallel_2023, gaidon_virtual_2016,tokmakov_learning_2021,nowruzi2019much}. The domain gap between real and synthetic data is recognized as a key problem and substantial effort has been dedicated to reducing the appearance gap~\cite{gaidon_virtual_2016,khirodkar_domain_2019}, with few notable papers addressing the content gap~\cite{prakash_structured_2019,haddad2021data,georgakis2017synthesizing}.
In most applications, the content gap is addressed through Domain Randomization~\cite{tremblay2018training,tobin_domain_2017}, where object placements are sampled from some broad, uninformative distribution, but we also discuss some more elaborate approaches below.
Overall, it has been demonstrated that training on synthetic data can produce state-of-the-art performance on real data~\cite{tokmakov_learning_2021,nowruzi2019much}, although it is typically advantageous to use a mix of synthetic and real data~\cite{nowruzi2019much}.

There is a major line of work addressing the content gap, and in particular the placement distribution, originating with Structured Domain Randomization (SDR)~\cite{prakash_structured_2019}, which used a highly structured, but hand-coded distribution to generate content for the scene. This approach was then extended to learn both the parameters~\cite{kar_meta-sim_2019} and the structure~\cite{vedaldi_meta-sim2_2020} of this distribution, eventually being able to learn the full distribution of both content and appearance from unlabelled real data~\cite{prakash_self-supervised_2021}. While some of those papers contain experiments similar to ours they do not isolate the impact of placement distribution, and their code and data are not available.

The basic approach for placing vehicles in a driving simulator, which we use as the baseline in this paper, is to spawn vehicles at designated locations and then allow the built-in behavioral models to drive them around and take a snapshot of their positions at some point in time. This produces limited variability due to the simplicity of the behavioral models and is often supplemented by Domain Randomization~\cite{tremblay2018training,tobin_domain_2017}, where the objects are placed in the scene at random according to some simple distribution. Scenes generated this way are often unrealistic, in particular in driving scenes many vehicles would be placed off-road. It is therefore common to manually engineer more complex distributions with domain-specific heuristics~\cite{sadeghi_cad2rl_2017}, which can perform well but require a lot of human effort. Another approach is to use ground truth object placements from real data and synthetically generate a variety of appearances~\cite{gaidon_virtual_2016}.

Among the learning-based approaches to object placement, SceneGen~\cite{tan_scenegen_2021} is a method specifically designed to learn the placement distribution of vehicles on the road and the paper contains an experiment that demonstrates how the realism of this distribution in synthetic training data impacts object detection on real data. The experiment is limited to LiDAR-based, rather than vision-based models, and the results are reported on a private dataset only. Various other models for placement distribution have been proposed, such as LayoutVAE~\cite{jyothi_layoutvae_2019} and Permutation Invariant Flows~\cite{zwartsenberg2023conditional}, but those papers do not study how using such models for synthetic data generation impacts downstream object detection performance. Since, to the best of our knowledge, none of those models are publicly available, we obtain our realistic placement samples by calling INITIALIZE, a public commercial API, which is learning-based but the details of the underlying model were not disclosed.

\section{Conclusion}
We have conducted a simple experiment that unambiguously shows that a realistic object placement distribution can have a dramatic impact on real-world performance when training object detectors on synthetic data in driving contexts. We believe that this placement distribution is a critical consideration when assembling synthetic datasets and that our paper will convince practitioners to pay close attention to this issue when working with synthetic data. To allow better reproducibility and comparisons with other placement generation methods, we make our code and datasets publicly available\footnote{\url{https://github.com/inverted-ai/object-detection}}.
\clearpage
{\small
\bibliographystyle{ieee_fullname}
\bibliography{inverted-ai,egbib}
}

\end{document}